\documentclass{article}

\PassOptionsToPackage{numbers,sort&compress}{natbib}
\usepackage[preprint]{neurips_2026}

\usepackage[utf8]{inputenc}
\usepackage[T1]{fontenc}
\usepackage{hyperref}
\usepackage{url}
\usepackage{booktabs}
\usepackage{amsfonts}
\usepackage{amsmath}
\usepackage{amssymb}
\usepackage{nicefrac}
\usepackage{microtype}
\usepackage{xcolor}
\usepackage{graphicx}
\usepackage{float}
\usepackage{multirow}
\usepackage{array}
\usepackage{bm}
\usepackage{enumitem}
\usepackage{titlesec}
\usepackage{fontawesome5}
\usepackage{subcaption}
\usepackage{wrapfig}

\titlespacing*{\paragraph}{0pt}{0.5ex plus 0.2ex minus 0.1ex}{0.6em}

\newcommand{\modelname}[1]{Modality Forcing{#1}}
\newcommand{\flux}[1]{FLUX.2-klein-9B{#1}}

\usepackage[table]{xcolor}
\definecolor{rankone}{RGB}{210,245,218}   %
\definecolor{ranktwo}{RGB}{255,242,204}   %
\definecolor{rankthree}{RGB}{219,235,255} %

\usepackage{multirow}
\usepackage{caption}      %
\usepackage{placeins}    %

\title{\modelname{} for Scalable Spatial Generation}

\author{%
  Bardienus Pieter Duisterhof$^{1,2}$ \quad
  Deva Ramanan$^{1}$ \quad
  Jeffrey Ichnowski$^{1}$ \\
  Justin Johnson$^{2}$ \quad
  Keunhong Park$^{2}$ \\[3pt]
  \normalfont $^1$Carnegie Mellon University \quad $^2$World Labs \\
   \faGlobe~\href{https://modality-forcing.github.io}{\texttt{Project Page}} 
  \quad
  \faGithub~\href{https://github.com/Duisterhof/modality-forcing}{\texttt{Code}} 
  \quad
  \raisebox{-0.2ex}{\includegraphics[height=0.95em]{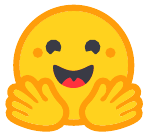}}~\href{https://huggingface.co/spaces/bartduis/modality_forcing}{\texttt{HuggingFace Demo}}
}

\begin{document}

\maketitle

\begin{figure}[H]
  \vspace{-10mm}
  \centering
  \begin{subfigure}{\linewidth}
    \centering
    \includegraphics[width=\linewidth]{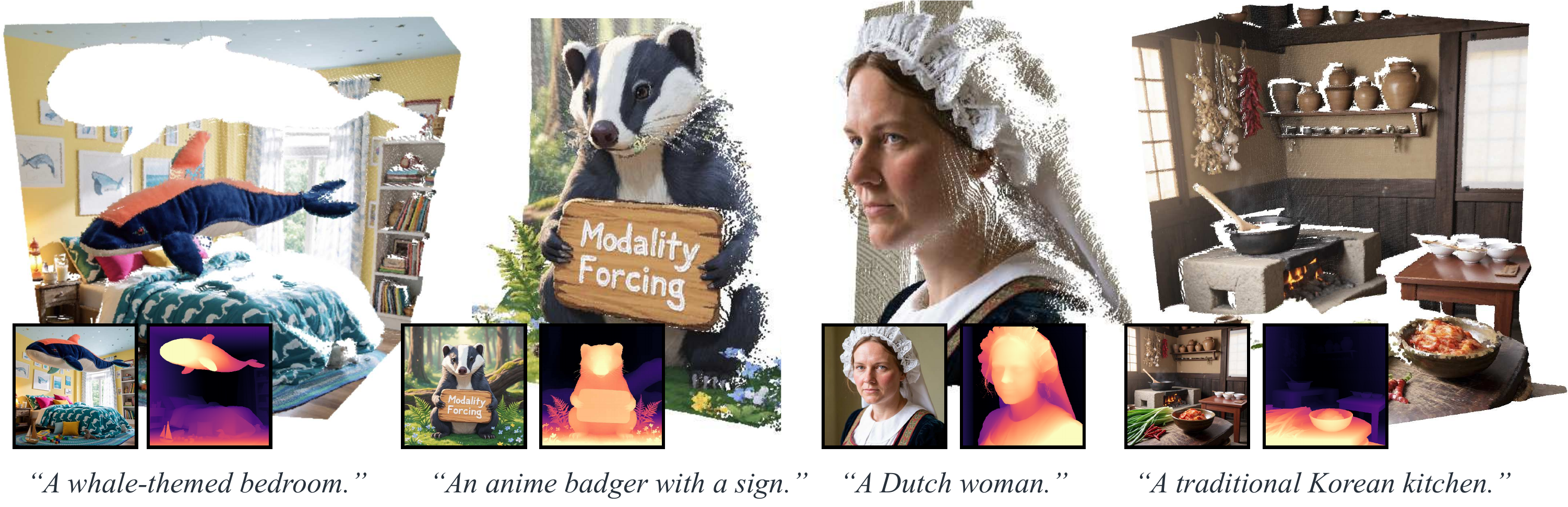}
    \caption{Joint Generation: Text $\Rightarrow$ Image + Depth}
    \label{fig:teaser_a}
  \end{subfigure}

  \vspace{1mm}

  \begin{subfigure}{0.59\linewidth}
    \centering
    \includegraphics[width=\linewidth]{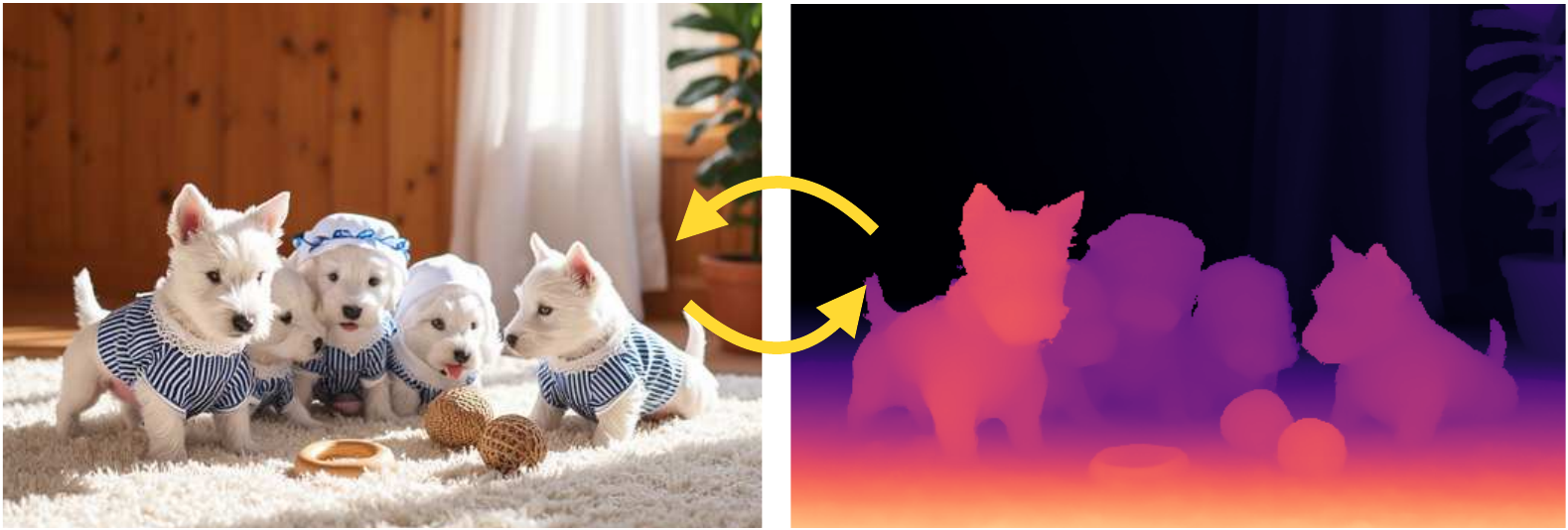}
    \caption{Conditional Generation: Image $\rightleftharpoons$ Depth}
    \label{fig:teaser_b}
  \end{subfigure}
  \hfill
  \begin{subfigure}{0.40\linewidth}
    \centering
    \includegraphics[width=\linewidth]{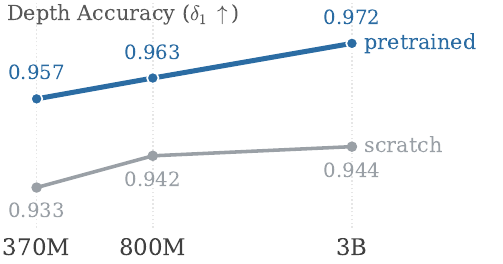}
    \caption{Scales with T2I model size}
    \label{fig:teaser_c}
  \end{subfigure}
  \caption{We present \modelname{}, a post-training recipe to extract spatial priors from text-to-image (T2I) models.
  A single DiT models the joint distribution over images and depth, enabling joint and conditional generation in arbitrary combinations. We demonstrate depth predictions improve with increasingly capable T2I pretraining, suggesting T2I is a scalable objective for spatial generation.}
  \label{fig:figure_1}
\end{figure}

\vspace{-7mm}
\begin{abstract}
Text-to-image (T2I) models contain rich spatial priors.
Synthesizing photorealistic, cluttered scenes requires an understanding of geometry, including perspective and relative scale.
Prior works adapt T2I models to leverage this prior for depth prediction, but they require dense depth data and involve complex recipes.
We propose \modelname{}, a simple, scalable post-training recipe for joint image-depth generation using a single DiT trained on sparse depth data.
\modelname{} enables conditional and joint generation of image and depth in any permutation by assigning separate noise levels per modality.
Per-modality decoders let us train on sparse, real-world depth and achieve strong, generalizable depth prediction.
We further show that \modelname{} inherits the scalability of T2I pre-training: by training a set of T2I models from scratch (370M to 3.3B parameters), we find that larger models trained on more image data produce more accurate depth.
Our strongest model is competitive with state-of-the-art monocular depth estimators and reduces AbsRel by 57\% relative to existing joint image-depth generative models.
These results provide strong evidence that image generation is a scalable pre-training objective for spatial perception.

\vspace{-12pt}%
\end{abstract}

\begin{figure*}[t]
    \centering
    \includegraphics[width=\textwidth]{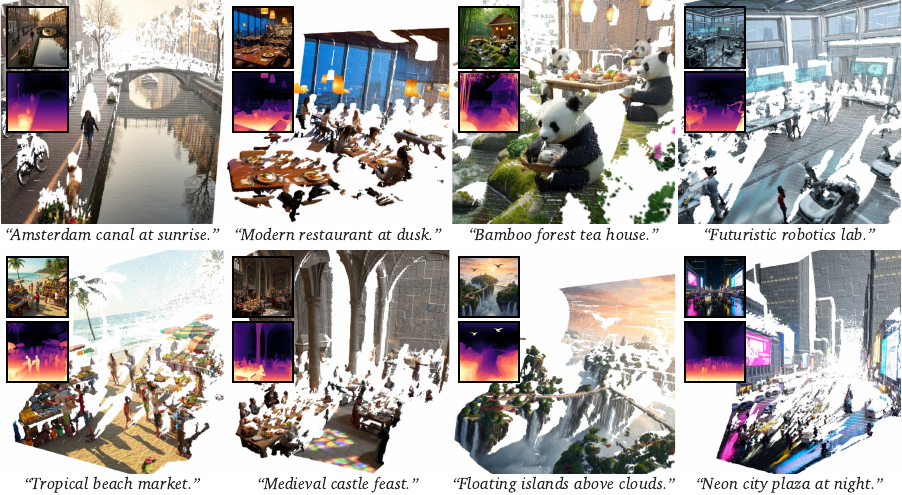}
    \caption{\modelname{} generates rich RGB-Depth from text prompts. Unprojecting the points to 3D, \modelname{} generates faithful and sharp geometry. The same checkpoint enables monocular depth estimation, and depth-to-image generation competitive with the best specialist models. }
    \label{fig:joint_gen_ours_only}
\end{figure*}
\section{Introduction}

Over the past several years, simple architectures and formulations fed with plentiful data have outperformed complex human-designed pipelines in many tasks.
Examples include DUSt3R~\cite{wang2024dust3rgeometric3dvision} for 3D reconstruction, SAM~\cite{kirillov2023segany} for segmentation, and SAM-3D~\cite{sam3dteam2025sam3d3dfyimages} for mesh generation.
A central lesson is that general methods capable of absorbing more data are more effective than complex heuristics---often called the \emph{bitter lesson}~\cite{Sutton2019BitterLesson}.

Models trained on task-specific data have led to considerable advances; however, 3D data is scarce which makes it difficult to scale to the quantities required for large models.
One approach to ease this is to combine multiple synergetic tasks in a unified multi-task formulation.
Language modeling is the clearest example, where a single generalist model trained on next-token prediction~\cite{DBLP:journals/corr/abs-2005-14165} exceeds specialist systems across translation, question answering, and coding.

Text-to-image (T2I) models contain similarly rich representations, but adapting them to new spatial tasks is difficult.
First, they denoise latent tokens in an image VAE which is not designed for arbitrary spatial modalities.
Depth as images limits training to dense (usually synthetic) samples and precludes encoding other spatial modalities such as meshes or point clouds. 
Second, post-training T2I models for spatial tasks may destroy its pretrained backbone and limit the pre-training benefit. 
Post-training generative models to produce spatial modalities requires a simple and scalable recipe.

We introduce \modelname{}, a post-training recipe that enables a pre-trained T2I model to predict depth.
\modelname{} combines latent RGB tokens and pixel-space depth tokens into a single DiT to model the joint image-depth distribution. Pixel-space depth diffusion enables learning from sparse real-world depth annotations and achieves state-of-the-art results. Once trained, \modelname{} allows setting RGB or depth as a target or conditioning, enabling joint generation, or conditional image-to-depth (I2D) or depth-to-image (D2I) generation.

Prior work has also modeled the joint RGB-depth distribution, but relied on learned adapters and was limited to dense depth data~\cite{zhang2023jointnetextendingtexttoimagediffusion,Byung-Ki_2025_ICCV}.
We show that post-training a single DiT on the joint distribution unlocks training on real-world data and improves depth predictions by 57\% on average.
Concurrent work explored native multimodal pre-training~\cite{yang2026contextunrollingomnimodels} and instruction fine-tuning~\cite{visionbanana2026}, representing depth as images.
These models show encouraging results across visual tasks but share the same challenges with data, and leave it unclear whether depth quality scales with T2I capability.

We contribute a controlled scaling study to investigate the scalability of \modelname{} with T2I model capability.
We train a family of DiT's from scratch, spanning from 370M to 3.3B parameters, and apply the same \modelname{} recipe to each.
We find that increasingly capable T2I models, larger models trained on more images, produce more accurate depth predictions. This suggests T2I is a scalable pre-training objective for spatial generation.

\paragraph{Contributions.}
\begin{itemize}[leftmargin=*, nosep]
    \item \textbf{Modality Forcing}, a simple post-training recipe that unifies three spatial tasks---monocular depth estimation (I2D), depth-to-image (D2I), and joint image-depth generation with one set of weights.
    \item A \textbf{controlled scaling study} from 370M to 3.3B parameters and from zero to 1.92B images, showing that depth predictions scale with T2I model and training size.
    \item \textbf{State-of-the-art results}: applied to \flux{}, \modelname{} competes with the best depth estimators and outperforms joint image-depth baselines by a margin of 57\%.
\end{itemize}

\section{Related Work}
\label{sec:related_work}

\paragraph{Text-to-image.}
Text-to-image (T2I) models have advanced rapidly over the past few years~\cite{rombach2022highresolutionimagesynthesislatent,esser2024scalingrectifiedflowtransformers,flux-2-2025,team2025zimage}.
The Stable Diffusion family of models began with latent diffusion models~\cite{rombach2022highresolutionimagesynthesislatent}, which perform denoising in the latent space of a pretrained VAE rather than in pixel space; this became the dominant template for efficient image synthesis.
A major architectural shift came with Stable Diffusion 3~\cite{esser2024scalingrectifiedflowtransformers}, which replaced the U-Net with a Diffusion Transformer (DiT)~\cite{peebles2023scalablediffusionmodelstransformers} trained under a rectified-flow objective.
The current state of the art for open-source models is largely set by FLUX~\cite{flux-2-2025} and Z-Image~\cite{team2025zimage} — both flow transformers operating in the latent space of an image VAE and trained on large volumes of curated data.
Whether this growing T2I capability reflects a correspondingly stronger underlying \emph{spatial} prior remains untested; \modelname{} answers this by post-training a controlled family of T2I checkpoints into joint image-depth generators under one recipe.

\paragraph{Monocular depth estimation.}
Monocular depth estimation has recently been transformed by foundation models trained on large corpora~\cite{depth_anything_v1,depth_anything_v2,wang2025moge,wang2025moge2}.
Depth Anything V2~\cite{depth_anything_v2} scales by distilling a large teacher trained on synthetic data through pseudo-labeled real images, yielding strong zero-shot depth.
Other works repurpose T2I models as depth priors: Marigold~\cite{ke2023repurposing} fine-tunes Stable Diffusion on synthetic RGB-D pairs.
The DiT architecture itself has also been adapted for depth: Pixel-Perfect Depth~\cite{xu2025pixel} shows that diffusing depth in pixel space avoids edge artifacts.
Most recently, MoGe~\cite{wang2025moge} reframes the problem as direct prediction of an affine-invariant 3D point map from which depth and intrinsics can be derived, and MoGe-2~\cite{wang2025moge2} extends this to metric scale.
Together, these lines of work establish that foundation-style learning on large 3D datasets yields high-quality depth and that T2I models carry exploitable spatial priors.
Despite the rapid progress of T2I models, the extent to which their spatial priors transfer to depth has not been systematically studied.

\paragraph{Joint Image-Depth Generation.}
A growing body of work tackles jointly generating images and depth, motivated by the observation that a single joint distribution over RGB and depth unifies conditional and joint generation.
LDM3D~\cite{stan2023ldm3dlatentdiffusionmodel} retrains a Stable Diffusion VAE and U-Net on six-channel RGB-D inputs.
JointNet~\cite{zhang2023jointnetextendingtexttoimagediffusion} duplicates the pretrained U-Net into a parallel depth branch tightly coupled to a frozen RGB branch, supporting bidirectional prediction via channel-wise inpainting.
UniCon~\cite{li2024unicon} generalizes this with LoRA and disentangled per-branch noise schedules, recovering depth-conditioned generation, depth estimation, and joint sampling in a single model.
JointDiT~\cite{Byung-Ki_2025_ICCV} ports the paradigm to diffusion transformers atop FLUX.1, introducing unbalanced per-modality timestep sampling.
Orchid~\cite{krishnan2025orchidimagelatentdiffusion} instead trains a joint VAE over color, depth, and normals, paired with a single latent diffusion model.
All of these methods rely on dense RGB-D supervision, typically from synthetic data or curated captured datasets. In contrast, \modelname{} is a post-training recipe that scales with the base T2I model and removes the requirement for dense supervision.

\begin{figure*}[t]
    \centering
    \begin{subfigure}[b]{0.75\linewidth}
        \centering
        \includegraphics[width=\linewidth]{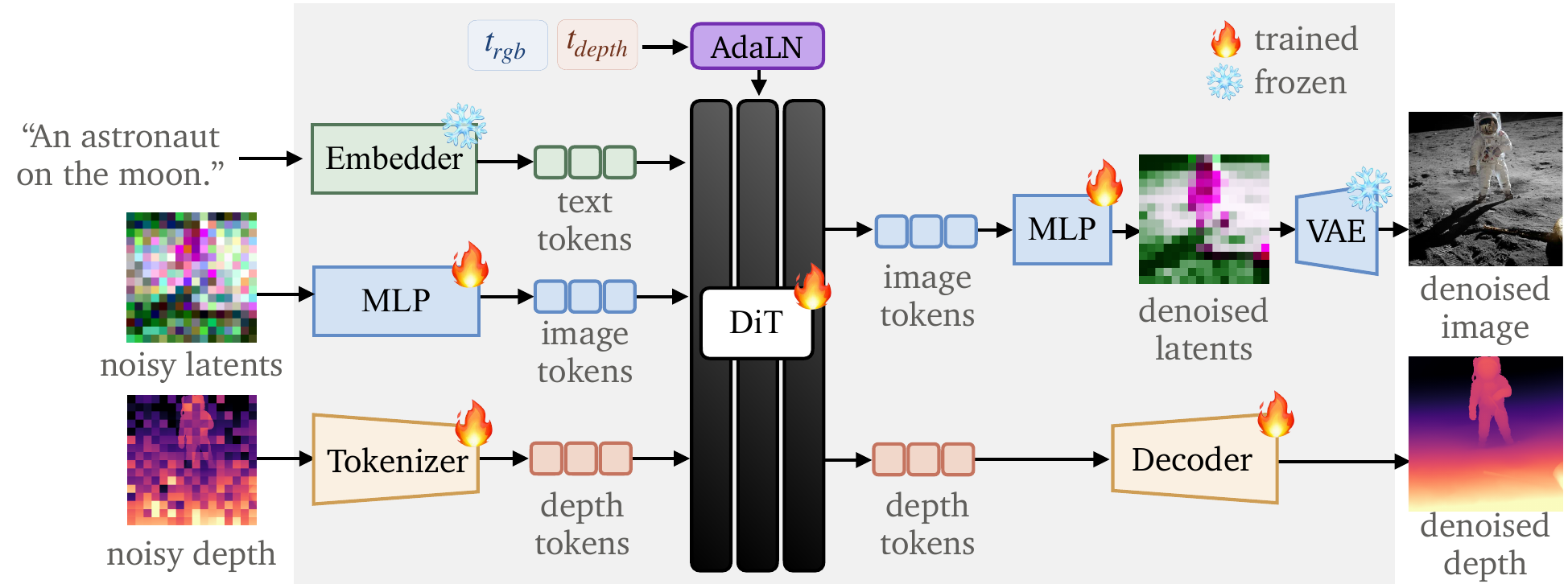}
        \caption{Training}
        \label{fig:method_training}
    \end{subfigure}
    \hfill
    \begin{subfigure}[b]{0.24\linewidth}
        \centering
        \includegraphics[width=\linewidth]{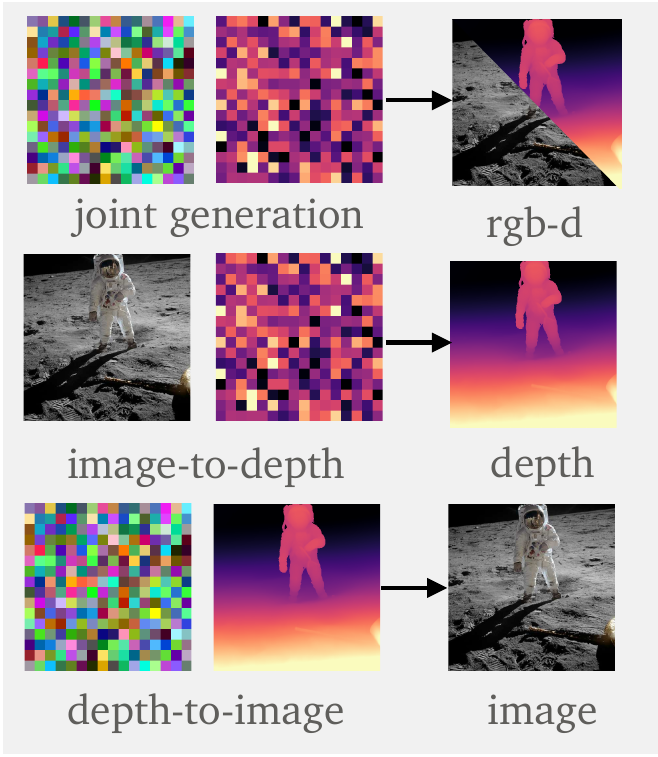}
        \caption{Inference}
        \label{fig:method_inference}
    \end{subfigure}
    \caption{
    \modelname{} is a recipe to post-train image-generation models for depth prediction. We encode RGB using a pretrained VAE and depth using a per-pixel tokenizer. 
    During training~(\subref{fig:method_training}), each modality is independently noised and trained with separate losses. 
    At inference~(\subref{fig:method_inference}), each modality's noise level can be controlled to yield image\textrightarrow{}depth, depth\textrightarrow{}image, or joint generation.
    }
    \label{fig:method}
\end{figure*}

\section{Method}
Modality Forcing post-trains a pretrained text-to-image DiT to model the joint distribution over RGB and depth. The core idea is to give each modality its own noise level, so that a single set of weights supports joint generation and either conditional direction. Tokenizing depth in pixel space further lets the model learn from sparse, real-world depth supervision.
\subsection{Problem Definition}
\label{sec:problem}

We unify three generation tasks under a single model: joint image and depth generation, depth to image, and image to depth.
Let $\mathbf{c}$ denote a text prompt, $\mathbf{x} \in \mathbb{R}^{H \times W \times 3}$ an RGB image, and $\mathbf{d} \in \mathbb{R}^{H \times W}$ a depth map.
We learn a joint generator
\begin{equation}
    p_\theta(\mathbf{x}, \mathbf{d} \mid \mathbf{c})
\end{equation}
from a set of $(\mathbf{c}, \mathbf{x}, \mathbf{d})$ triples.
A single model $p_\theta$ then supports three tasks during inference:
\begin{align}
     (\mathbf x, \mathbf d) & \sim p_\theta(\mathbf x, \mathbf d \mid \mathbf c) && \text{Joint RGB-D generation} \\
     \mathbf d & \sim p_\theta (\mathbf d \mid \mathbf x, \mathbf c) && \text{Image-to-depth (I2D)} \\
     \mathbf x & \sim p_\theta( \mathbf x \mid \mathbf d, \mathbf c ) && \text{Depth-to-image (D2I)}
\end{align}

\paragraph{Preliminaries.}
Diffusion models are an expressive class of generative models for high-quality generation. 
We discuss flow-matching with $v$-prediction~\cite{lipman2023flowmatchinggenerativemodeling} and $x$-prediction~\cite{li2026basicsletdenoisinggenerative}.
Let $\mathbf{x}_0 \sim q(\mathbf{x})$ denote a clean data distribution and $\boldsymbol{\epsilon} \sim \mathcal{N}(\mathbf{0}, \mathbf{I})$ Gaussian noise.
We interpolate linearly between the two with a scalar $t \in [0, 1]$ to create training samples,
\begin{equation}
    \mathbf{x}_t = (1 - t)\,\mathbf{x}_0 + t\,\boldsymbol{\epsilon},
\end{equation}
so that $\mathbf{x}_0$ is clean and $\mathbf{x}_1$ is pure noise.
The constant velocity along this path is $\mathbf{v} = \mathbf{x}_0 - \boldsymbol{\epsilon} $.
With \emph{v-prediction} we parameterize the output as the velocity and regress a network $g_\theta(\mathbf{x}_t, t)$ directly:
\begin{equation}
    \mathcal{L}_{\text{v-pred}} = \mathbb{E}\big[\lVert g_\theta(\mathbf{x}_t, t) - \mathbf{v}\rVert_2^2\big].
\end{equation}
Recent work has shown that $v$-prediction can struggle in high-dimensional spaces~\cite{li2026basicsletdenoisinggenerative}, instead suggesting to parameterize the output as the clean sample, $\hat{\mathbf{x}}_0 = g_\theta(\mathbf{x}_t, t)$.
The implied velocity $\hat{\mathbf{v}} = (\hat{\mathbf{x}}_0 - \mathbf{x}_t ) / t$ is regressed to $\mathbf{v}$:
\begin{equation}
    \mathcal{L}_{\text{x-pred}} = \mathbb{E}\!\left[\left\lVert\big(g_\theta(\mathbf{x}_t, t)- \mathbf{x}_t  \big)/t - \mathbf{v}\right\rVert_2^2\right].
\end{equation}
This is well-conditioned when the data lies on a low-dimensional manifold (such as natural images or depth maps).
Both methods sample the learned vector field from $t = 1$ to $t = 0$ with an ODE solver.

\subsection{Modality Forcing}
\label{sec:modality_forcing}
We introduce \modelname{}, a diffusion algorithm with per-modality noise levels.
Assigning separate noise levels allows conditional and joint generation in any permutation: an image with no noise conditions, while a full noise image is a generation target. Several methods use this idea, but differ in the axis along which the noise varies. Teacher Forcing~\cite{6795228} runs along sequence position, holding past tokens clean while denoising the next. Diffusion Forcing~\cite{chen2025diffusion} varies noise level per token. Latent Forcing~\cite{baade2026latentforcingreorderingdiffusion} denoises DINO~\cite{oquab2023dinov2} latents ahead of raw pixels. \modelname{} extends this idea to the axis of modality: RGB and depth carry their own noise levels within a diffusion process, so that fixing a modality at $t=0$ makes it conditioning. The schedule alone selects joint, image-to-depth, or depth-to-image generation.

We post-train existing T2I models to support all sampling schemes.
During training, we sample per-modality noise levels in three different ways to support our three generation tasks.
Joint RGB-D generation samples $t_{\text{rgb}},t_{\text{depth}} \in [0,1]$;
I2D fixes $t_{\text{rgb}} = 0$ and samples $t_{\text{depth}} \in [0,1]$;
and D2I samples $t_{\text{rgb}} \in [0, 1]$ and fixes $t_{\text{depth}} = 0$.

\paragraph{Depth tokenizer.} Real-world videos typically only contain sparse depth annotation, since depth is estimated, e.g., using multi-view stereo (MVS) pipelines. To allow for post-training on these sparse annotations, we denoise depth directly in pixel space rather than the latent VAE space. %

This is in contrast to prior work~\cite{Byung-Ki_2025_ICCV, li2024unicon, ke2023repurposing} which tokenizes depth through a pre-existing or new image VAE, with no obvious mechanism to accommodate partial depth supervision. %
We fill missing pixels with isotropic Gaussian noise to signal to the model that depth is not available at those locations, equivalent to how a fully missing depth map is encoded. 

\paragraph{Per-modality timestep conditioning.}
Because each modality is denoised at its own noise level, every token must be modulated by its own timestep. We give RGB and depth separate timestep embedders: the RGB stream reuses the pretrained embedder, while depth receives a freshly initialized one. Since joint and conditional generation couple the two noise levels, we additionally let each stream's modulation observe the other modality's timestep through a lightweight cross-stream mixing module---one small embedder per direction. We initialize both to zero, so the embedding begins as an exact identity with no cross-communication and learns the coupling over training.

\paragraph{Depth detokenizer.}
The DiT blocks are followed by a depth detokenizer with $n$ layers of self-attention, followed by a final linear layer that maps the depth tokens back to pixel space.
The extra depth blocks help create depth-specific capacity without disrupting the RGB stream. We normalize depth supervision by scaling it to have unit mean, followed by spatial contraction~\cite{barron2022mipnerf360} to ensure $d \in [0,2]$.

\paragraph{Initialization.}
We warm-start the depth pathway rather than training it from scratch. We initialize the depth stream by cloning the pretrained image-stream weights; the remaining depth-specific modules---the pixel-space tokenizer, the depth timestep embedder, the detokenizer, and the cross-stream mixers---are initialized from scratch.

\subsection{Self-Distillation}
\label{sec:self_dist}
T2I models are trained on billions of images; post-training them on millions of less diverse images may erode the rich prior. To preserve it, we introduce a self-distillation loss that penalizes the student for drifting from the original T2I checkpoint (similar in spirit to `Learning without Forgetting'~\cite{li2017learningforgetting}). At train time, we pass the current noisy RGB through the frozen T2I model and record its predicted velocity $\mathbf{v}_{\text{t2i}}$, then penalize the deviation of the student's RGB velocity $\mathbf{v}$:
\begin{equation}
    \mathcal{L}_{\text{dist}} = (\lambda_{\text{hi}}t_{\text{depth}} + \lambda_{\text{lo}}(1-t_{\text{depth}}))||\mathbf{v}_{\text{t2i}} - \mathbf{v}||_2^2,
\end{equation}
where $||\mathbf{v}_{\text{t2i}} - \mathbf{v}||_2^2$ is the L2 loss between the predicted velocity and the T2I-predicted velocity. We set $\lambda_{\text{hi}} > \lambda_{\text{lo}}$ so the penalty is strongest at $t_{\text{depth}}=1$: when depth is fully noised it carries no information. As depth is denoised toward $t_{\text{depth}}=0$, it supplies context the T2I model never sees, so we relax the penalty rather than force agreement with a now-uninformed teacher.

\begin{figure*}[t]
    \centering
    \begin{minipage}[c]{0.49\linewidth}
        \centering
        \scriptsize
        \captionof{table}{\textbf{Training data.} We train on
        twelve real-world and simulated datasets totaling 17M frames across
        58k scenes, spanning indoor scanning, outdoor driving, and synthetic rendering.}
        \label{tab:training_data}
        \label{tab:training_data}
\centering
\begin{tabular}{l r r}
\toprule
Dataset & \# Scenes & \# Frames \\
\midrule
Argoverse 2          & 700     & 108{,}139    \\
Aria Project Sim     & 5{,}022 & 2{,}424{,}158 \\
ARKitScenes          & 5{,}041 & 744{,}200    \\
Blended MVS          & 493     & 113{,}209    \\
FoundationStereo     & 40{,}936 & 40{,}936    \\
Hypersim             & 457     & 74{,}519     \\
MegaDepth            & 113     & 19{,}447     \\
ParallelDomain       & 1{,}520 & 239{,}840    \\
ScanNet v2           & 1{,}200 & 94{,}751     \\
TartanAir v2         & 1{,}122 & 8{,}563{,}146 \\
Taskonomy            & 520     & 4{,}385{,}534 \\
Waymo Open           & 796     & 159{,}000    \\
\midrule
\textbf{Total}       & \textbf{57{,}920} & \textbf{16{,}966{,}879} \\
\bottomrule
\end{tabular}

        \vspace{1em}
        \captionof{table}{Base T2I model sizes used for scaling experiment, ranging from 370M to 3.3B parameters.}
        \label{tab:t2i_models_scratch}
\centering
\begin{tabular}{l r r r}
\toprule
Param Count  & Token Size & FFN Dim & Depth \\
\midrule
370M   & 1024 & 3,072  & 30 \\
800M   & 1536 & 4,608  & 30 \\
3.3B & 3072 & 9,216  & 30 \\
\bottomrule
\end{tabular}

    \end{minipage}
    \hfill
    \begin{minipage}[c]{0.47\linewidth}
        \centering
        \includegraphics[width=\linewidth]{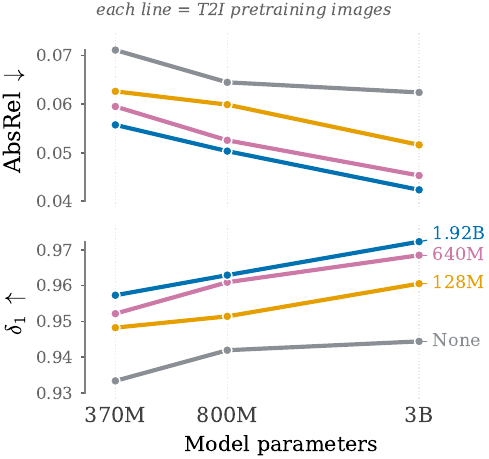}
        \captionof{figure}{Scaling experiments. Depth accuracy ($\delta_1$, $\uparrow$, bottom) and AbsRel ($\downarrow$, top) by T2I model size. Each line represents a T2I pre-training dataset size (none, 128M, 640M, 1.92B). Training larger T2I models on more image data yields better depth performance.}
        \label{fig:scaling}
    \end{minipage}
\end{figure*}
\section{Results}
\label{sec:experiments}
We evaluate \modelname{} across joint and conditional RGB-Depth tasks. First, we train a suite of T2I models from scratch to study how depth generation scales with T2I model and training dataset size. Next, we apply \modelname{} to \flux{}, and benchmark it against the best specialist models. Our experiments show that depth quality scales reliably with T2I model size and data, and that \modelname{} is competitive with the best depth models.

\subsection{Scaling experiment.}
\label{sec:scaling}
We conduct a controlled scaling experiment to answer: \emph{does depth performance scale with T2I capability?}
We train three T2I models on a large and diverse set of web images at 256$\times$256 resolution and post-train with \modelname{}.
We evaluate depth performance across model and training dataset sizes.
There should be improvements across both axes if our hypothesis is true.

Each T2I model is a DiT trained with a flow-matching objective.
For each training pair $(\mathbf{x}, \mathbf{c})$, we encode the image into a clean latent $\mathbf{z}_0$ with the frozen FLUX.2 VAE~\cite{flux-2-2025}.
We sample a timestep $t \in [0, 1]$ and add noise to the latent along the linear interpolant $\mathbf{z}_t = (1-t)\,\mathbf{z}_0 + t\,\boldsymbol{\epsilon}$ with $\boldsymbol{\epsilon} \sim \mathcal{N}(\mathbf{0}, \mathbf{I})$.
The network output $g_\theta(\mathbf{z}_t, t, \mathbf{c})$ is parameterized as a clean-latent prediction $\hat{\mathbf{z}}_0$, and we regress the implied velocity to the true velocity $\mathbf{v} = \mathbf{z}_0 - \epsilon$:
\begin{equation}
    \mathcal{L}_{\text{T2I}} = \big\lVert (g_\theta(\mathbf{z}_t, t, \mathbf{c}) - \mathbf{z}_t)/t - \mathbf{v} \big\rVert_2^2.
\end{equation}
The resulting checkpoints are the starting points for \modelname{} post-training.

We vary model size from 370M to 3.3B (Table~\ref{tab:t2i_models_scratch}), and T2I pre-training data size from zero to 1.92B samples.
For depth, we use $\approx$17M training frames, a combination of real and synthetic images (Table~\ref{tab:training_data}), and evaluate depth across NYUv2~\cite{nyu_v2_dataset}, DIODE~\cite{diode_dataset}, ETH3D~\cite{eth3d_dataset} and ScanNet~\cite{scannet_dataset}.

Figure~\ref{fig:scaling} shows depth performance across model and pre-training data size. The results suggest that the performance of \modelname{} scales with T2I model size and training data. Larger T2I models and larger training data result in improved depth quality. This suggests \modelname{} is a scalable recipe for depth generation. The results show that T2I pre-training and not just model size enables better depth quality, presenting direct evidence of the spatial prior present in T2I models.

\paragraph{Implementation Details}
The T2I trunk is a flat-token DiT~\cite{peebles2023scalablediffusionmodelstransformers} with parallel attention and MLP branches per block, pre- and post-RMSNorm, grouped-query attention, and RoPE positional encoding~\cite{su2023roformerenhancedtransformerrotary}.
Latents are produced by a frozen FLUX.2 VAE~\cite{flux-2-2025} and tokenized by a fresh patch embedding; text prompts are encoded by a frozen UMT5-XXL~\cite{chung2023unimaxfairereffectivelanguage} into embeddings, and the trunk attends to these text tokens at every block. We use logit-normal sampling for RGB, and plateau logit-normal sampling for depth, both with a timestep shift~\cite{esser2024scalingrectifiedflowtransformers} of $\mu=1.1$.
Additionally, we set $p_{\text{i2d}} = 0.2$ and $p_{\text{d2i}}=0.2$, to focus the network on the case of a fully denoised image or depth map.

Captioning follows protocols similar to those in SD3~\cite{esser2024scalingrectifiedflowtransformers} and WAN~\cite{wan2025wanopenadvancedlargescale}.
Each model in the family has the same number of layers but a different token size (Table~\ref{tab:t2i_models_scratch}).

The metrics are averaged across NYUv2~\cite{nyu_v2_dataset}, DIODE~\cite{diode_dataset}, ETH3D~\cite{eth3d_dataset} and ScanNet~\cite{scannet_dataset} datasets and take 256 $\times$ 256 center crops. We use the robust affine-invariant alignment introduced in MoGe-2~\cite{wang2025moge2}.

\begin{figure*}[t]
    \centering
    \includegraphics[width=0.96\textwidth]{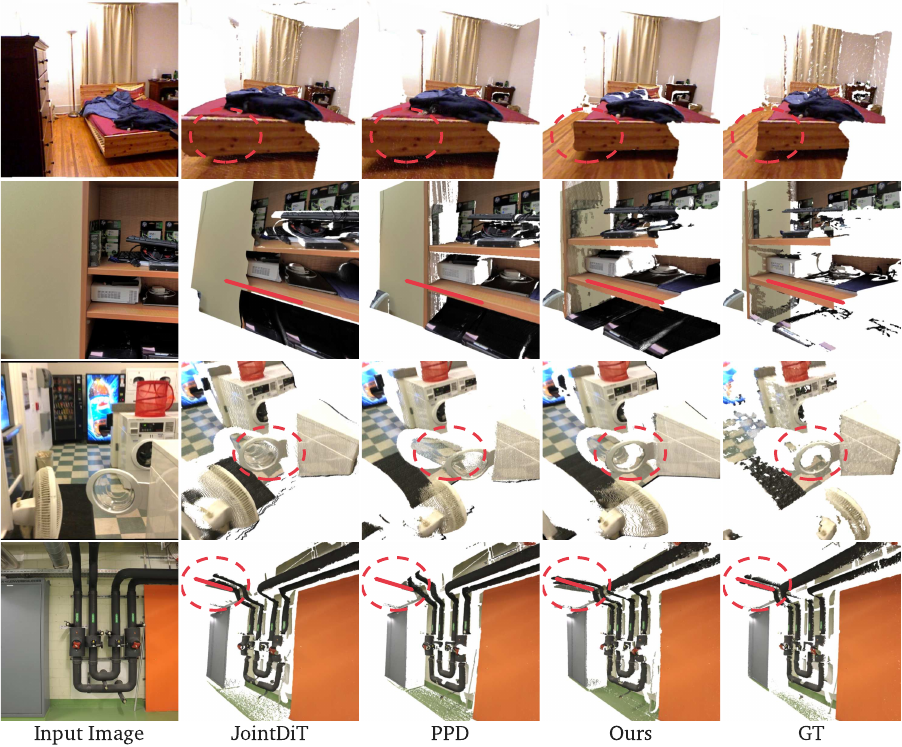}
    \caption{Qualitative image-to-depth generation results.
    \modelname{} generates robust and sharp depth maps yielding plausible 3D point clouds.
    We find JointDiT~\cite{Byung-Ki_2025_ICCV} sometimes fails catastrophically, missing structure or misestimating scale.
    PPD~\cite{xu2025pixel} produces much closer results, but \modelname{} produces more robust results.
    }
    \label{fig:i2d}
    \vspace{-3mm}
\end{figure*}

\subsection{FLUX-Based \modelname{}}
\begin{table*}[t]
\vspace{-3mm}
\caption{\textbf{Affine-invariant depth estimation.}
Comparison of monocular depth estimation methods on five benchmarks.
The results suggest \modelname{} outperforms all models built for joint RGB-Depth generation, and generative depth estimators. 
\modelname{} approaches the performance of the very best depth models such as MoGe-2~\cite{wang2025moge2}, scoring better on NYUv2~\cite{nyu_v2_dataset} and ETH3D~\cite{eth3d_dataset}. The overall best are \underline{underlined}, within-type best are \textbf{bolded}.
}
\label{tab:zero_shot_depth}
\centering
\scriptsize
\setlength{\tabcolsep}{3.5pt}
\resizebox{\textwidth}{!}{%
\begin{tabular}{l l cc cc cc cc cc}
\toprule
& & \multicolumn{2}{c}{NYUv2~\cite{nyu_v2_dataset}}
  & \multicolumn{2}{c}{KITTI~\cite{kitti_dataset}}
  & \multicolumn{2}{c}{ETH3D~\cite{eth3d_dataset}}
  & \multicolumn{2}{c}{ScanNet~\cite{scannet_dataset}}
  & \multicolumn{2}{c}{DIODE~\cite{diode_dataset}} \\
\cmidrule(lr){3-4}\cmidrule(lr){5-6}\cmidrule(lr){7-8}\cmidrule(lr){9-10}\cmidrule(lr){11-12}
Method Type & Method
  & AbsRel$\downarrow$ & $\delta_1\uparrow$
  & AbsRel$\downarrow$ & $\delta_1\uparrow$
  & AbsRel$\downarrow$ & $\delta_1\uparrow$
  & AbsRel$\downarrow$ & $\delta_1\uparrow$
  & AbsRel$\downarrow$ & $\delta_1\uparrow$ \\
\midrule
\multirow{6}{*}{\shortstack[l]{Discriminative\\depth estimation}}
 & ZoeDepth~\cite{bhat2023zoedepthzeroshottransfercombining}        & 4.76 & 97.3 & 5.59 & 95.1 & 7.27 & 94.2 & --   & --   & 7.80 & 90.9 \\
 & MASt3R~\cite{leroy2024groundingimagematching3d}          & 4.67 & 96.7 & 5.79 & 95.1 & 4.64 & 97.0 & --   & --   & 5.79 & 94.1 \\
 & DA-v2~\cite{depth_anything_v2}          & 4.16 & 97.9 & 6.77 & 94.3 & 4.63 & 97.2 & --  & -- & 5.41 & 94.6 \\
 & Depth Pro~\cite{bochkovskii2025depthprosharpmonocular}       & 3.67 & 98.2 & 5.12 & 96.8 & 4.97 & 96.4 & --   & --   & 4.66 & 95.6 \\
 & UniDepth V2~\cite{piccinelli2024unidepth}     & 2.96 & \textbf{98.6} & 3.85 & \underline{\textbf{98.1}} & 2.95 & 98.5 & --   & --   & 4.05 & 96.5 \\
 & MoGe-2~\cite{wang2025moge2}         & \textbf{2.89} & \textbf{98.6} & \underline{\textbf{3.75}} & \underline{\textbf{98.1}} & \textbf{2.80} & \textbf{99.1} & --   & --   & \underline{\textbf{3.14}} & \textbf{97.4} \\
\midrule
\multirow{4}{*}{\shortstack[l]{Generative\\depth estimation}}
   & Marigold~\cite{ke2023repurposing}        & 4.88 & 96.8 & 9.05  & 90.3 & 4.90 & 97.2 & 5.89 & 95.2 & 6.13 & 94.5 \\
   & Lotus~\cite{he2025lotusdiffusionbasedvisualfoundation}           & 4.31 & 97.3 & 8.89  & 90.1 & 5.25 & 96.7 & 5.01 & 96.5 & 6.70 & 93.8 \\
   & GeoWizard~\cite{fu2024geowizardunleashingdiffusionpriors}       & 4.89 & 96.6 & 10.97 & 86.6 & 6.15 & 95.3 & 5.49 & 95.7 & 6.37 & 94.0 \\
   & PPD~\cite{xu2025pixel}             & \textbf{3.82} & \textbf{97.7} & \textbf{6.57}  & \textbf{95.1} & \textbf{4.02} & \textbf{98.3} & \textbf{4.04} & \textbf{97.8} & \textbf{4.97} & \textbf{95.6} \\
  \midrule
  \multirow{3}{*}{\shortstack[l]{I2D from\\Joint Model}}
   & JointNet~\cite{zhang2023jointnetextendingtexttoimagediffusion}        & 11.92 & 86.7 & 13.74 & 81.3 & 12.63 & 87.1 & 14.81 & 81.8 & 20.02 & 82.7 \\
   & UniCon~\cite{li2024unicon}          & 8.75  & 91.7 & 18.67 & 73.0 & 8.77 & 92.9 & 9.23 & 91.0 & 13.57 & 90.8 \\
   & JointDiT~\cite{Byung-Ki_2025_ICCV}        & 5.12  & 96.9 & 10.90 & 88.4 & 5.32 & 97.0 & 5.76 & 96.3 & 10.22 & 93.9 \\
& Ours           & \underline{\textbf{2.52}}    & \underline{\textbf{98.9}}    & \textbf{5.37}    & \textbf{96.6}    & \underline{\textbf{2.37}}    & \underline{\textbf{99.3}}    & \underline{\textbf{2.32}}    &
  \underline{\textbf{98.9}}    & \textbf{3.35}    & \underline{\textbf{97.7}}    \\
\bottomrule
\end{tabular}}
\end{table*}

\begin{figure*}[t]
    \centering
    \includegraphics[width=\textwidth]{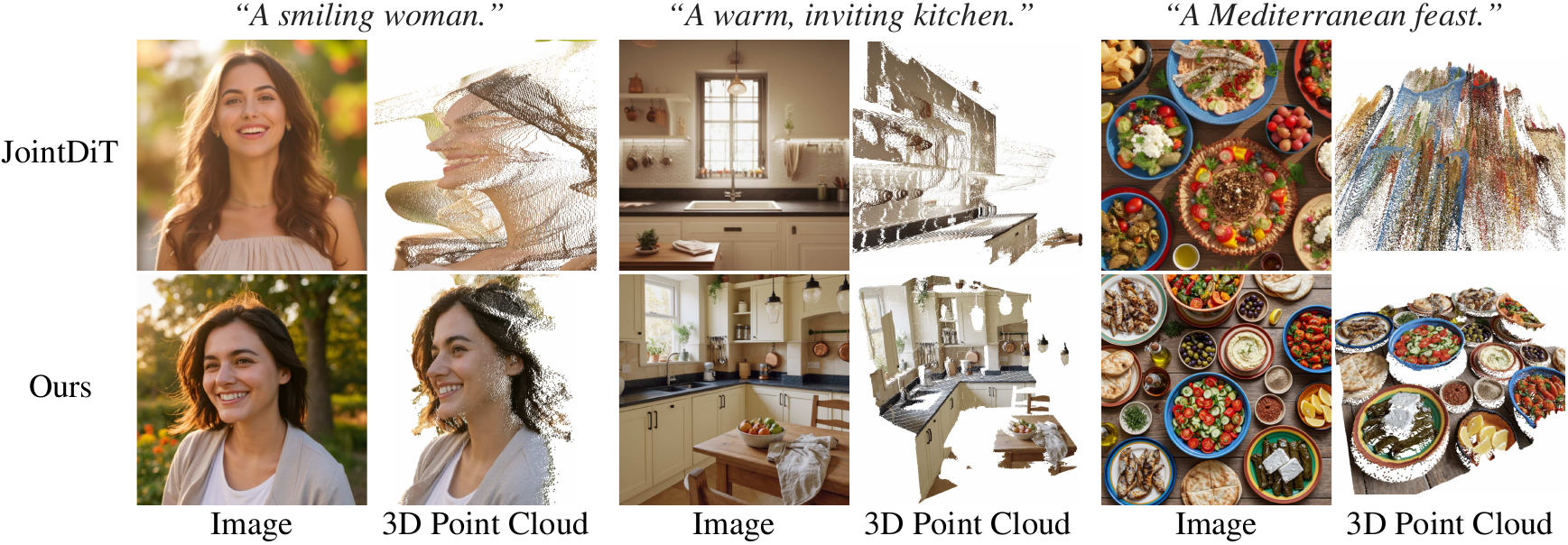}
    \caption{Qualitative joint image-depth generation results.
    \modelname{} samples RGB and geometry jointly, and proves to produce consistent and compelling point clouds. JointDiT~\cite{Byung-Ki_2025_ICCV} inherits compelling image generations from FLUX, but struggles with depth. }
    \label{fig:joint_gen}
    \vspace{-5mm}
\end{figure*}
After showing \modelname{} scales with T2I capability, we now evaluate its performance when paired with a SoA T2I model.
We post-train \flux{} with the \modelname{} recipe and evaluate it on conditional and joint generation tasks.
We post-train this model using the real-world and synthetic depth data shown in Table~\ref{tab:training_data}.
The training recipe is largely unchanged from Section~\ref{sec:scaling}, only now we apply the self-distillation loss to retain more of the T2I prior (Section~\ref{sec:self_dist}).

\paragraph{Image-to-depth generation}
We evaluate the monocular depth estimation capability by evaluating affine-invariant depth on NYUv2~\cite{nyu_v2_dataset}, DIODE~\cite{diode_dataset}, ETH3D~\cite{eth3d_dataset}, ScanNet~\cite{scannet_dataset} and KITTI~\cite{kitti_dataset}. 
We evaluate the \modelname{} recipe against state-of-the-art depth estimation models. 
The comparison includes discriminative depth estimators, generative depth models, and existing joint image-depth generators.
Discriminative depth estimators include the most performant models such as MoGe-2~\cite{wang2025moge2} and Depth Pro~\cite{bochkovskii2025depthprosharpmonocular}. 
We also include generative depth estimators such as pixel-perfect-depth (PPD)~\cite{xu2025pixel}.
Finally, most comparable to \modelname{}, we directly compare against methods which jointly generate image and depth such as JointDiT~\cite{Byung-Ki_2025_ICCV}.

The results in Table~\ref{tab:zero_shot_depth} suggest that \modelname{} convincingly outperforms existing joint image-depth models and generative depth models.
We attribute the performance delta to the scalability of our recipe to real-world depth data, the strong prior from the \flux{} backbone, and our post-training recipe.
\modelname{} is competitive even to the strongest depth models such as PPD~\cite{xu2025pixel}, Depth Anything V2~\cite{depth_anything_v2} and MoGe-2~\cite{wang2025moge2}.
\modelname{} outperforms all prior baselines on NYUv2~\cite{nyu_v2_dataset} and ETH3D~\cite{eth3d_dataset}. Our performance on ScanNet~\cite{scannet_dataset} is also strong, but confounded by the ScanNet training split in our data mixture.

Qualitatively, Figure~\ref{fig:i2d} shows that the predicted depth is consistent with ground truth.
We find that JointDiT~\cite{Byung-Ki_2025_ICCV} infers predictions that violate geometric constraints such as straight lines and vertical walls.
We also observe JointDiT's quantization artifacts for distant predictions. 

\paragraph{Joint Generation}
Joint image-depth generation is a key capability enabling prompt-conditioned scene generation for artists, architects and embodied agents.
Figure~\ref{fig:joint_gen} shows a comparison between \modelname{} and the existing SoA joint image-depth generator JointDiT~\cite{Byung-Ki_2025_ICCV}.
JointDiT produces stunning images, but the depth predictions are not as robust as the \modelname{} recipe.

\begin{wraptable}{r}{0.5\linewidth}
    \vspace{-\baselineskip}        %
    \centering
    \small                          %
    \caption{{Depth-conditioned image generation on 6,000 OpenImages samples. }
}
\label{tab:depth_to_image}
\centering
\begin{tabular}{l cc}
\toprule
& \multicolumn{2}{c}{OpenImages 6K} \\
\cmidrule(lr){2-3}
Method & FID$\downarrow$ & AbsRel$\downarrow$ \\
\midrule
Readout-Guidance  & 18.72 & 23.19 \\
ControlNet              & 13.68 & 9.85  \\
UniCon~\cite{li2024unicon}                      & 13.21 & 9.26  \\
JointDiT~\cite{Byung-Ki_2025_ICCV}                                    & 12.62 & \textbf{6.99} \\
\midrule
\modelname{} (Ours)                                     & \textbf{11.41} & 9.26  \\
\bottomrule
\end{tabular}
\end{wraptable}
\noindent\textbf{Depth-to-Image Generation (D2I)}
Depth-to-image generation allows for generating various samples from a single depth map, conditioned on a text prompt.
This unlocks generating various scenes with identical layout for embodied AI agents, or assets with identical geometry but different appearance.
We follow the evaluation protocol from JointDiT~\cite{Byung-Ki_2025_ICCV} and UniCon~\cite{li2024unicon} by evaluating on 6,000 images in the OpenImages dataset with Depth Anything V2~\cite{depth_anything_v2} depth annotations. We conduct two experiments: (1) generate all 6,000 images conditioned on DAv2 depth, and compute the FID vs GT images, (2) compute AbsRel between DAv2(generated images) and DAv2(GT images), this is a proxy for depth-following.

Table~\ref{tab:depth_to_image} shows the results. \modelname{} generates images with the lowest FID, considerably better than all the baselines. We find D2I performance does not match JointDiT~\cite{Byung-Ki_2025_ICCV}, as our recipe appears to more loosely follow depth instructions for some scenes.

\paragraph{Implementation Details}
Unlike other T2I models, FLUX.2 uses double-stream layers (an image and text stream). 
We append a depth stream for those layers and freeze the RGB weights for the double stream layers. 
We train this model at 512 $\times \, x$ variable aspect ratio, where $x \leq 512$. 

We use logit-normal sampling for RGB, and plateau logit-normal sampling for depth, both with a timestep shift of $\mu=1.1$.
Additionally, we set $p_{\text{i2d}} = 0.2$ and $p_{\text{d2i}}=0.2$, to focus the network on the case of a fully denoised image or depth map.

We use the robust and optimal alignment solver (ROE)~\cite{wang2025moge2}, as well as the ensemble approach documented by Marigold~\cite{ke2023repurposing} with $N=10$. 
We rerun the joint and generative models to match the protocols, we copy the discriminative results from the MoGe-2 paper as they match our protocol.

\subsection{Denoising Trajectory Ablations}

\modelname{} allows setting arbitrary per-modality noise levels at inference time, a flexibility we previously showed enables competitive image-to-depth and depth-to-image generation. Here we leverage the same mechanism to extend our analysis to partial depth conditioning and to the model's behavior across a range of different generation trajectories.

\begin{figure*}[htb!]
    \centering
    \begin{minipage}[t]{0.49\linewidth}
        \vspace{0pt}
        \noindent\textbf{Depth Quality.}
        Encoding per-modality timesteps independently allows for arbitrary denoising trajectories.
        We study how RGB and depth generations are affected by various trajectories through the noise landscape -- prioritizing image or depth tokens early on in the denoising process.
        We follow Latent Forcing~\cite{baade2026latentforcingreorderingdiffusion} by parameterizing the denoising trajectory according to $f_{\alpha}(t) = \frac{\alpha t}{1+(\alpha-1)t}$, with $t_{\text{depth}} = f_{\alpha}(t_{\text{rgb}})$. For $\alpha > 1$, we denoise RGB first, while $\alpha \in [0,1)$ means depth is prioritized first. We sweep $\alpha \in [2^{-5}, 2^{5}]$ and complete inference from the OpenImages 6k prompts. We then compute depth metrics between MoGe-2~\cite{wang2025moge2} on the generated image and our own generated depth map.
        The results suggest that denoising RGB first yields more consistent depth. This suggests that generating natively in the latent VAE space may be an easier space to traverse than raw depth pixels, similar to what was found in Latent Forcing~\cite{baade2026latentforcingreorderingdiffusion}.
    \end{minipage}
    \hfill
    \begin{minipage}[t]{0.49\linewidth}
        \vspace{0pt}
        \centering
        \begin{subfigure}[b]{\linewidth}
            \centering
            \includegraphics[width=0.9\linewidth]{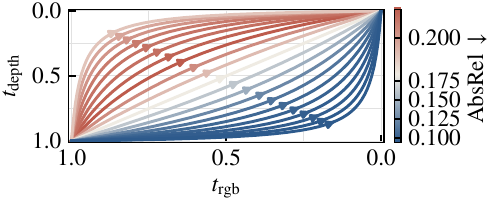}
            \vspace{-3mm}
            \caption{Inference-time trajectories studied.}
            \label{fig:inference_trajs}
        \end{subfigure}
        \hfill
        \begin{subfigure}[b]{\linewidth}
            \centering
            \includegraphics[width=0.9\linewidth]{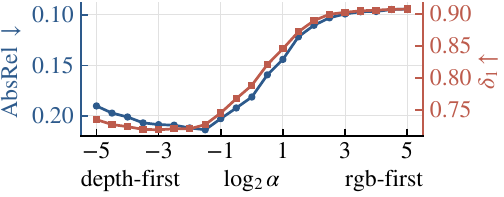}
            \caption{Denoising RGB first leads to better depth prediction.}
            \label{fig:method_inference}
        \end{subfigure}
        \caption{\modelname{} inference-time analysis. We find that denoising RGB first acts as a kind of `scratch pad' in latent space, leading to higher-quality depth predictions.}
        \label{fig:trajectories}
    \end{minipage}
\end{figure*}
\begin{figure}[t]
    \centering
    \includegraphics[width=\linewidth]{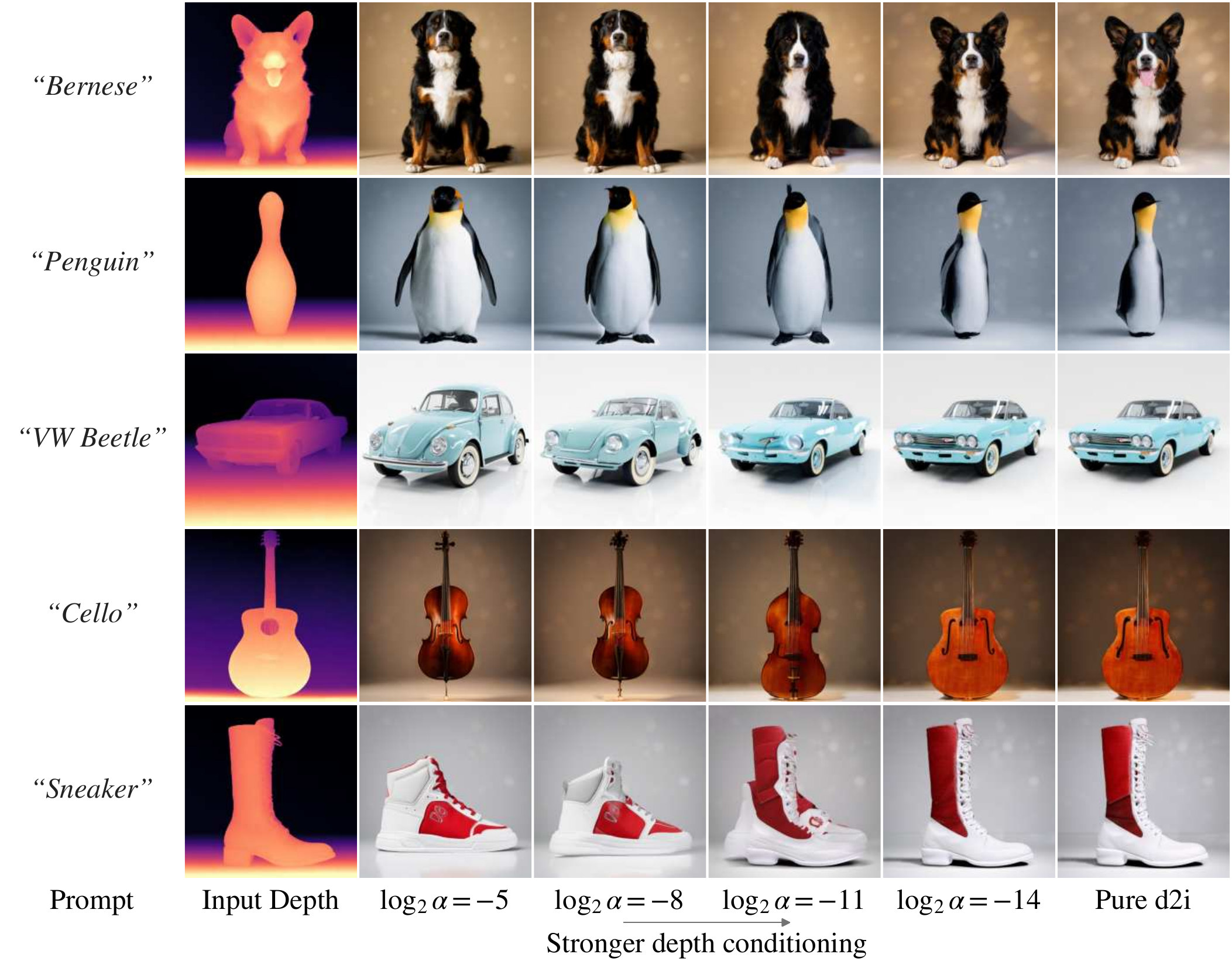}
    \caption{The denoising trajectory across depth and rgb dictates the strength of modality conditioning. Denoising more of depth early on means RGB will more rigidly match it. Here we show that this enables a new level of controllability of depth-conditioned image generation. }
    \label{fig:partial_depth_cond}
\end{figure}

\paragraph{Partial Depth Conditioning}
In some creative applications, a user may want to control the conditioning strength between modalities — using a depth map to loosely fix coarse object pose, or to dictate precise geometry. Arbitrary denoising trajectories enable this partial conditioning: at every step we set $\mathbf{x}_{\text{depth}} = (1-t_{\text{depth}})\mathbf{x}_{\text{cond}} + t_{\text{depth}}\epsilon$. 
Denoising depth earlier means more of RGB generation happens with clean depth, yielding stronger conditioning. Figure~\ref{fig:partial_depth_cond} shows that this mechanism effectively controls depth conditioning strength, and yields compelling and viable results. This simple test-time implementation adds another capability to the same model.

\section{Conclusion and Limitations}
\label{sec:conclusion}
We present \modelname{}, a simple post-training recipe that turns a T2I model into a unified image-depth generator.
The central idea is to assign each modality its own timestep and loss while sharing a single DiT backbone, which enables joint and conditional generation in any combination.
Across controlled scaling experiments, the results suggest that stronger T2I pretraining transfers to stronger spatial prediction.
Our strongest model based on \flux{} achieves I2D performance competitive with the very best depth models.
This supports the broader view that web-scale generative image models contain spatial structure that can be exposed and continues to scale.

The current model also leaves clear directions for improvement in future work.
First, our scaling study is limited to at most 3B parameters and does not derive a full scaling law. 
Future work may explore scaling to larger T2I models and study scaling behavior in greater detail.
Second, our experiments use relatively few depth samples, several orders of magnitude fewer than what is seen during T2I pretraining.
Following recent work~\cite{wang2026vggtomega}, we will likely see another leap in performance by scaling up depth data and the T2I backbone to > 9B parameters.
Finally, architectural adjustments may further reduce artifacts or enable prediction of metric depth. 
With these extensions, we believe \modelname{} provides a broad foundation for scalable generation across modalities.
Applied to depth, \modelname{} competes with the very best specialist models and is shown to continue to scale.

\section{Acknowledgements}

We thank Gengshan Yang, Katja Schwarz, Ben Mildenhall, Hao Zhang, Andy Cheng and other colleagues for productive discussions during the project and in revising the manuscript.

\bibliographystyle{plainnat}
\bibliography{main}

\end{document}